\documentclass[10pt]{article}
\usepackage[a4paper,margin=1in]{geometry}
\usepackage[T1]{fontenc}
\usepackage{lmodern}
\usepackage{microtype}
\usepackage{amsmath,amssymb,amsthm}
\usepackage{booktabs}
\usepackage{multirow}
\usepackage{graphicx}
\usepackage{xcolor}
\usepackage[numbers,sort&compress]{natbib}
\usepackage[colorlinks=true,linkcolor=blue!50!black,citecolor=blue!50!black,urlcolor=blue!50!black]{hyperref}
\usepackage[capitalise,noabbrev]{cleveref}
\usepackage{caption}
\newtheorem{proposition}{Proposition}
\newtheorem{lemma}{Lemma}
\theoremstyle{definition}

\newcommand{\R}{\mathbb{R}}
\newcommand{\tol}{\mathrm{tol}}
\newcommand{\Th}{\Theta}
\newcommand{\nTh}{\lVert\Theta\rVert}
\newcommand{\best}[1]{\textbf{#1}}
\newcommand{\second}[1]{\underline{#1}}

\newif\ifanon
\anonfalse
\ifanon
  \newcommand{\codeurl}{\url{https://anonymous.4open.science/r/rolewalk}}
\else
  \newcommand{\codeurl}{\url{https://github.com/kerighan/rolewalk}}
\fi

\title{\textbf{RoleWalk: training-free structural embeddings\\ for directed graphs that transfer across graphs}}
\author{\ifanon Anonymous authors\else Maixent Chenebaux\\ {\normalsize Trickstr}\fi}
\date{}

\begin{document}
\maketitle

\begin{abstract}
Structural role embeddings describe a node by its position in a graph --- hub, bridge, periphery, member of a dense core --- rather than by who its neighbours are.
Existing methods either learn coordinates specific to one graph (struc2vec, Role2Vec, RESD) or read scales and normalisations from the whole graph (GraphWave, Digraphwave), so the same neighbourhood can receive different coordinates in two graphs; few read edge directions.
We present \emph{RoleWalk}, a training-free embedding that follows a short lazy random walk from each node and records, at each step, two distributions: that of the transition probabilities, through its characteristic function at a fixed grid of frequencies, and that of the walker's distance to its starting node.
On directed graphs, walks that go forward then backward add who shares a node's targets or sources.
The embedding depends only on a node's neighbourhood within a fixed radius, so its coordinates mean the same thing in every graph.
We show that, without curvature, it is the kernel mean embedding of the walk's footprint, that it separates graphs that the Weisfeiler--Leman test cannot, and that pruning small probabilities comes with a per-node error certificate.
Empirically, RoleWalk is competitive with the best role embeddings on five classification benchmarks at a fraction of their cost, has the highest scores on real directed roles in Wikipedia adminship, exceeds the published accuracy of Digraphwave at every noise level of its own Enron network-alignment benchmark, transfers centrality predictors to graphs four times larger better than existing inductive embeddings, hand-made features and a trained GIN, and embeds a million nodes in 20\,s to 3\,min on one CPU core.
The code is open source (BSD license): \codeurl.
\end{abstract}

\section{Introduction}

\begin{figure}[t]
\centering
\includegraphics[width=\linewidth]{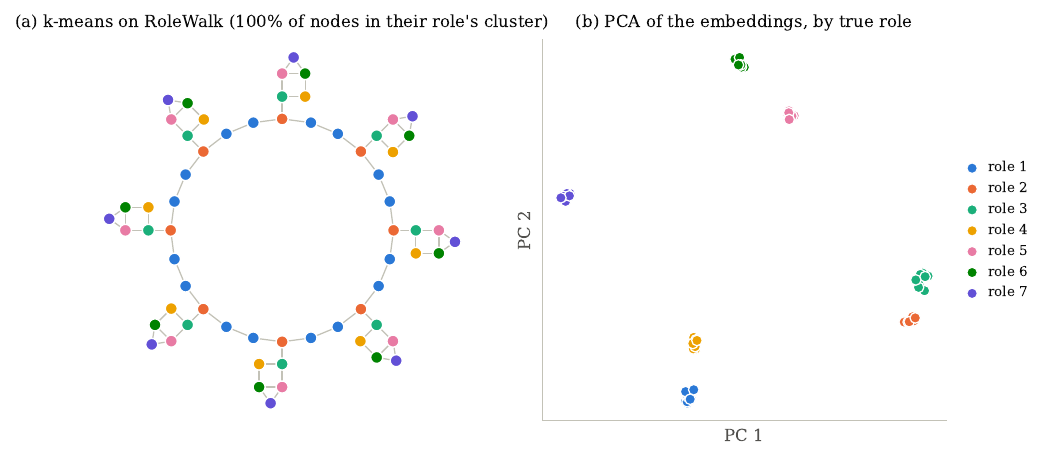}
\caption{Structural roles on eight houses attached to a cycle, the setting of GraphWave \citep{donnat2018}. (a) Nodes coloured by $k$-means ($k = 7$) on default RoleWalk embeddings; colours are matched to the seven true roles, and every node lands in its role's cluster. Far-apart copies of the same position get the same coordinates. (b) First two principal components of the standardised embeddings, coloured by true role; nodes in the same role coincide exactly, and a small jitter is added to show them.}
\label{fig:roles}
\end{figure}

Two airports can play the same role --- a regional hub feeding a national one --- without sharing a single route.
\emph{Structural} embeddings capture this kind of similarity: nodes are close when their neighbourhoods look alike, wherever they are in the graph \citep{henderson2012,ribeiro2017,donnat2018}.
They are used for role discovery, for network alignment, where a node must be recognised in a second graph \citep{heimann2018,ceylan2022}, and as positional or structural encodings in graph neural networks and graph transformers \citep{dwivedi2022,rampasek2022}.

Three properties matter in these uses, and current methods rarely combine them.
\emph{Transfer:} for alignment, or for a model trained on one graph and applied to another, a coordinate must mean the same thing in every graph.
Methods that fit a skip-gram model or an autoencoder per graph \citep{ribeiro2017,ahmed2018,resd} produce coordinates that are only defined up to that fit; GraphWave \citep{donnat2018} averages over all $n$ nodes and picks its diffusion scale from each graph's spectrum, so identical neighbourhoods in graphs of different sizes get different coordinates.
\emph{Directions:} in a vote, a citation or an email, who points to whom is part of the role, yet most role embeddings drop directions.
\emph{Cost:} in our benchmarks, learned and diffusion methods take seconds to minutes on graphs of a thousand nodes, and most do not run at a hundred thousand within an hour or 16\,GB.

RoleWalk starts from a simple observation.
After $w$ steps of a random walk from node $u$, two things characterise $u$'s position: how \emph{concentrated} the walk is (a hub spreads its mass thinly, a leaf keeps most of it) and how \emph{far} it has gone (a node in a triangle-rich region keeps returning near its start).
RoleWalk encodes both, at each of a few steps, with fixed functions that need no training and no graph-level quantity.

\paragraph{Contributions.}
\begin{itemize}
\item \textbf{A local, training-free embedding} built from two laws of a short lazy walk (\cref{sec:method}), with \emph{mixed probes} that let walks change direction on directed graphs. It is deterministic, reads weights, and has 69 coordinates on undirected graphs and 178 on directed ones.
\item \textbf{Properties} (\cref{sec:theory}): locality and invariance, an interpretation as a kernel mean embedding, triangle counting and incomparability with 1-WL, a provable failure class, and a per-node certificate for pruned computation.
\item \textbf{Experiments} (\cref{sec:experiments}): role classification on five networks, real directed roles, network alignment on Digraphwave's own protocol, transfer of centralities across graph sizes, and scaling to a million nodes.
\end{itemize}
RoleWalk is an open-source Python library (BSD license; \codeurl), released with the benchmark scripts and every result file; a complete technical report with all ablations, negative results and additional benchmarks accompanies it as supplementary material.

\section{Related work}

\paragraph{Structural roles.}
ReFeX \citep{henderson2011} aggregates degree-like features recursively over neighbours, and RolX \citep{henderson2012} factorises them into roles.
struc2vec \citep{ribeiro2017} builds a multilayer graph of degree-sequence similarities and learns skip-gram coordinates; struc2vec++ \citep{steenfatt2018} extends it to directed and weighted graphs.
Role2Vec \citep{ahmed2018} and RiWalk \citep{ma2019riwalk} relabel nodes by structural signatures and run skip-gram on walks over the relabelled graph, and RESD \citep{resd} learns a variational autoencoder on structural features; the RONE survey \citep{rone} benchmarks these methods.
xNetMF, used in REGAL \citep{heimann2018}, embeds neighbourhood degree distributions against landmarks fitted on the graphs to align.
Learned coordinates can be shared across graphs by freezing an encoder or a reference basis; RoleWalk's coordinates are instead computed independently for each node, with no fitted component.

\paragraph{Characteristic functions and diffusion signatures.}
GraphWave \citep{donnat2018} embeds each node by the empirical characteristic function of its column of the heat kernel.
RoleWalk uses the same summary on a different object --- the rows of powers of a lazy walk --- and drops GraphWave's two graph-level quantities, the $1/n$ average and the spectral choice of scale (\cref{prop:graphwave} in the appendix).
Digraphwave \citep{ceylan2022} extends GraphWave to directed graphs with diffusion on the graph and its transpose and aggregation over neighbours.
FEATHER \citep{rozemberczki2020feather} also uses random-walk transition probabilities and characteristic functions, but in the opposite role: it takes the characteristic function of node \emph{features} (such as log-degree or clustering) at the nodes where the walk lands, weighted by the transition probabilities, whereas RoleWalk takes the characteristic function of the probabilities themselves and needs no node feature.

\paragraph{Walks, distances and directions.}
Anonymous walks \citep{ivanov2018anonymous} describe a walk by the order in which it first visits nodes, which, like RoleWalk's layers, records how far and how often a walk returns, but are sampled and used for graph-level embeddings.
Distance encodings \citep{li2020distance} give GNNs the shortest-path distance or landing probabilities between a target node set and every node; RoleWalk's layer block is the law of the shortest-path distance under the walk distribution, a summary of such an encoding that has a fixed size per node.
The mixed probes are normalised, lazy forms of bibliographic coupling \citep{kessler1963} (sharing targets) and co-citation \citep{small1973} (sharing sources), used here inside a structural embedding.

\paragraph{Structural encodings and expressiveness.}
Random-walk structural encodings (RWSE) \citep{dwivedi2022} use the return probabilities of the simple walk and are standard in GPS \citep{rampasek2022}; RoleWalk's distance-0 coordinate is the lazy-walk analogue (\cref{app:proofs}).
We compare expressiveness with the 1-WL test \citep{weisfeiler1968,xu2019,morris2019} and on BREC \citep{wang2023}.

\paragraph{Positioning.}
Each ingredient of RoleWalk has precedents: characteristic functions of diffusion rows, return probabilities, distances under a walk, coupling and co-citation.
What is new is their combination into fixed, local, training-free coordinates for directed and weighted graphs, with a kernel reading of the whole and certified pruning; and the empirical finding that its components matter for different tasks: the walker's distance law carries the transfer of centralities across graph sizes, while the characteristic function and the mixed probes carry network alignment.

\section{Method}\label{sec:method}

\begin{figure}[t]
\centering
\includegraphics[width=\linewidth]{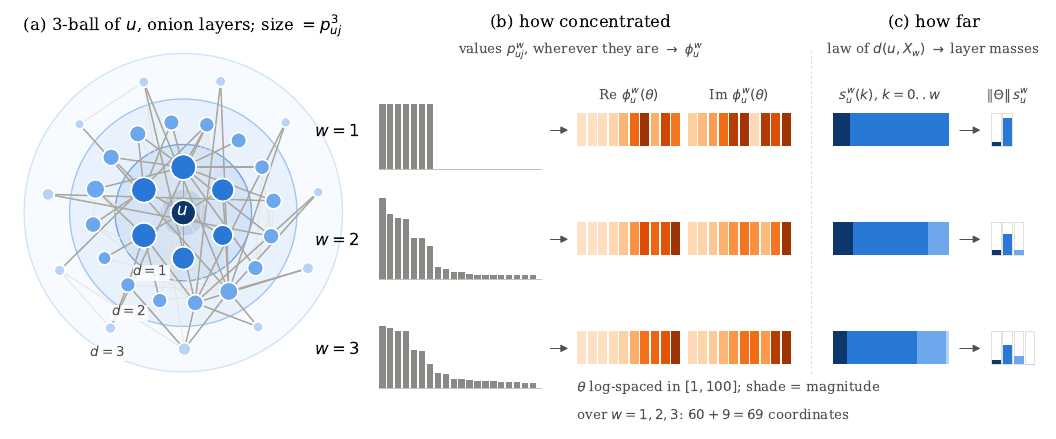}
\caption{RoleWalk on one node $u$ of Zachary's karate club. (a) The three-step walk from $u$ never leaves the ball of radius 3, drawn by onion layer (distance to $u$); node size is the probability of the walker being there after three steps. At each step $w$, RoleWalk reads two laws of the same walk, in parallel. (b) The multiset of its probabilities (grey: their location is ignored) gives the real and imaginary parts of the characteristic function $\phi^w_u$ at ten frequencies (shade: magnitude, scaled per block). (c) The law of its distance to $u$, the mass in each onion layer, gives $w+1$ coordinates scaled by $\nTh$.}
\label{fig:method}
\end{figure}

\paragraph{Lazy walk.}
Let $G=(V,E)$ be a graph with non-negative weights $A$, directed or not.
Self-loops in the input are ignored. For node $u$, let $s_u$ be its weighted out-degree and $d_u$ its number of out-neighbours with positive weight.
The walk stays put with probability $1/(d_u+1)$ and otherwise moves along an edge in proportion to its weight:
\begin{equation}\label{eq:walk}
H_{uu} = \frac{1}{d_u+1}, \qquad H_{uv} = \frac{d_u}{d_u+1}\,\frac{A_{uv}}{s_u} \quad (v \ne u),
\end{equation}
and a node without out-neighbours is absorbing ($H_{uu} = 1$).
On unweighted graphs this is $H = \tilde D^{-1}(I+A)$, the walk with one self-loop per node; on weighted graphs the self-loop weighs an average edge, so $H$ does not depend on the unit of the weights.
We write $p^w_u = e_u^\top H^w$ for the distribution of the walker after $w$ steps from $u$.

\paragraph{How concentrated: the characteristic function.}
With $f_\theta(x) = e^{i\theta x} - 1$ and a fixed grid $\Th = \{\theta_1,\dots,\theta_m\}$ ($m=10$ frequencies log-spaced in $[1,100]$),
\begin{equation}\label{eq:phi}
\phi^w_u(\theta) = \sum_{j\in V} f_\theta\big(p^w_{uj}\big), \qquad \theta \in \Th.
\end{equation}
Since $f_\theta(0)=0$, only the nodes the walk reaches contribute.
With $\Th$ an interval, $\phi^w_u$ determines the multiset of nonzero transition probabilities exactly (\cref{app:proofs}); ten frequencies spread over two decades sample it at the scales where probabilities live.

\paragraph{How far: the law of the distance.}
Let $d(u,j)$ be the shortest-path distance in the direction of the walk (out-distance for the out-walk, in-distance for the in-walk) and $X_w$ the walker's position after $w$ steps. The second law is
\begin{equation}\label{eq:layers}
s^w_u(k) = \Pr\big(d(u,X_w) = k\big) = \sum_{j:\,d(u,j)=k} p^w_{uj}, \qquad k = 0,\dots,w .
\end{equation}
It takes $w+1$ values, and the multiset of probabilities ignores it: it says whether the walk lingers near $u$ (loops, triangles) or moves out.
We call the sets of nodes at each distance \emph{onion layers} (\cref{fig:method}). The distances come from the supports of earlier powers, so no extra walk is computed.

\paragraph{Embedding.}
For each step $w = 1,\dots,L$ ($L = 3$ by default), RoleWalk concatenates $\operatorname{Re}\phi^w_u$, $\operatorname{Im}\phi^w_u$ and $\nTh\, s^w_u$, where $\nTh = (\sum_\theta \theta^2)^{1/2}$.
The factor $\nTh$ is not tuned: it is the scale at which the layer block matches, to first order in $\theta p$, one characteristic function per layer, so that both blocks read as one kernel (\cref{prop:mmd}); without it the layers carry 0.02\% of the variance of raw embeddings. Since our classifiers standardise features, this scaling matters for raw distances ($k$-means, nearest neighbours, alignment), not for the classification tables.
This gives $2mL + \sum_{w\le L}(w+1) = 60 + 9 = 69$ coordinates.

\paragraph{Directed graphs: walks that change direction.}
On a directed graph we run the out-walk $H_+$ (on $A$) and the in-walk $H_-$ (on $A^\top$) and concatenate both embeddings.
These see where a node sends and where it receives, but not \emph{who shares its targets} --- two voters who back the same candidates, two predators of the same prey.
We therefore add the characteristic functions of the rows of $H_+H_-$ (one step forward, one back) and $H_-H_+$, the \emph{mixed probes}, for 178 coordinates in all.
\Cref{prop:mixed} shows a pair of nodes that only the mixed probes separate.

\paragraph{Optional curvature.}
On the simple unweighted symmetrisation of the graph, for an edge $xy$, $\kappa(x,y) = \sum_z \min(m_x(z), m_y(z))$ with $m_x$ the one-step lazy walk from $x$, which upper-bounds Ollivier--Ricci curvature \citep{ollivier2009,jost2014} and equals $(c_{xy}+2)/(\max(d_x,d_y)+1)$ with $c_{xy}$ the number of common neighbours.
Each node gets the mean $\kappa_x$ over its edges ($\kappa_x = 0$ at isolated nodes), and each step adds $\nTh$ times the mean and standard deviation of $\kappa$ under $p^w_u$ (75 coordinates undirected, 190 directed).

\paragraph{Computation and pruning.}
For each root, a single pass propagates a sparse row with a dense accumulator and summarises it without storing the matrix powers: memory is $O(n)$ per thread and rows are independent.
Entries below $\tol = 10^{-4}$ are dropped after each product, except at the first step; the dropped mass $r$ is added back to $\operatorname{Im}\phi$ to first order ($\theta r$). Distances are read from supports before pruning, which keeps them exact up to three steps; beyond, a node first reached through a pruned entry can be assigned a later layer.
Pruning is accurate on average but can be badly wrong on hubs, whose mass is spread over many tiny entries.
With a \emph{mass budget} $\delta$, a row drops small entries only while the total mass it has lost stays below $\delta$.
The library returns, for every node, the lost mass and, when the conditions of \cref{prop:prune} hold, the resulting error certificate.

\paragraph{Hyperparameters.}
Unless stated otherwise, experiments use the defaults ($L=3$, $m=10$, $\Th$ log-spaced in $[1,100]$, $\tol=10^{-4}$). Centrality transfer uses four-step RoleWalk with exact walks ($\tol=0$); its dimension-matched top-$k$ ablation uses four steps pruned at $10^{-5}$. The four steps were fixed when the centrality benchmark was first written and have not been varied since. Expressiveness evaluations use exact walks, and the pruning and walk-length experiments vary these settings explicitly.
The defaults were chosen on a grid of 48 configurations (frequency weighting, range, number of steps, layers) scored by unsupervised $k$-means agreement, with equal weight on synthetic motifs and on the airport, Reality-call and Actor networks; they remain the best whichever dataset is left out.
These networks are also those of the classification table (with a different metric); wiki-RfA, network alignment, centrality transfer, BREC and ZINC played no part in the choice.

\section{Properties}\label{sec:theory}

Throughout, $B_r(u)$ is the subgraph induced by the nodes within distance $r$ of $u$ (in the underlying undirected graph on directed graphs and with curvature), and statements hold without pruning ($\tol=0$), in exact arithmetic. Proofs are in \cref{app:proofs}.

\begin{proposition}[Locality and invariance]\label{prop:local}
The embedding of $u$ is determined by the isomorphism class of the rooted graph $(B_r(u),u)$, with $r=L$ for walks on undirected graphs, $r=\max(L,2)$ on directed graphs with mixed probes, and $r=L+2$ with curvature.
In particular it does not depend on the size of the graph, and nodes related by an automorphism get equal embeddings.
\end{proposition}

This is what makes transfer well defined: a coordinate means the same thing in every graph. The finite walk length is essential: removing GraphWave's $1/n$ factor and spectral scale alone would not make heat diffusion local.

\begin{proposition}[One kernel on the walk's footprint]\label{prop:mmd}
Mark each node $j$ reached at step $w$ by $(w,\allowbreak p^w_{uj},\allowbreak d(u,j))$, and let $\mu_u$ be the resulting finite measure (the \emph{footprint} of the walks from $u$).
Without curvature, $\lVert X_u - X_v\rVert^2$ is the squared maximum mean discrepancy \citep{gretton2012} between $\mu_u$ and $\mu_v$ for the positive semidefinite kernel
\begin{multline*}
K\big((w,a,\ell),(w',b,\ell')\big) \\ = \delta_{ww'}\Big[\textstyle\sum_{\theta\in\Th}\big(\cos\theta(a-b) - \cos\theta a - \cos\theta b + 1\big) + \nTh^2\, ab\,\delta_{\ell\ell'}\Big].
\end{multline*}
The first term compares probabilities whatever their layer, the second compares masses in the same layer; the second is the first-order part of the first restricted to one layer, and $\nTh$ is the only positive factor for which the two agree to first order. On directed graphs the step mark becomes a block mark (direction and step, or mixed probe), and mixed-probe blocks have no layer term.
\end{proposition}

This gives the two blocks of RoleWalk a common interpretation as one kernel comparison of the same object, rather than a unique prescription. With a finite grid the distance is an MMD pseudometric: the kernel is not proved characteristic.

\begin{proposition}[Expressiveness]\label{prop:expr}
Let $G$ be simple, undirected and unweighted.
\textbf{(a)} If $G$ is $d$-regular, $s^2_u(1) = (2d+2t_u)/(d+1)^2$, with $t_u$ the number of triangles at $u$. Hence RoleWalk with $L\ge 2$ separates graphs that 1-WL cannot, such as two triangles and a hexagon, or any two $d$-regular graphs with different triangle counts.
\textbf{(b)} Conversely, without curvature and for $L\in\{2,3\}$, there are nodes that $L$-round 1-WL separates and RoleWalk with $L$ steps does not, for every choice of frequencies: the two are incomparable.
\textbf{(c)} Distance-regular graphs with the same intersection array (for instance strongly regular graphs with equal parameters) get identical embeddings at all nodes, for every $L$, with or without curvature.
\end{proposition}

The counterexamples in (b) have 15 and 16 nodes: on all 6,781 nodes of the 996 connected graphs with at most 7 nodes, RoleWalk with 3 steps confuses no pair that 3-round 1-WL separates, and splits 571 of the 1-WL classes.
Part (c) is a sharp failure class; RWSE fails on it for the same reason.

\begin{proposition}[Walks that change direction]\label{prop:mixed}
Let $u$ be a source pointing to two sinks, or one of the two sources of a directed $K_{2,2}$. For every $L$, with or without curvature, $u$ gets the same out- and in-walk embedding in both graphs; the mixed probe $e_u^\top H_+H_-$ has values $(2/3,1/6,1/6)$ in the first and $(5/9,2/9,1/9,1/9)$ in the second, and separates them.
\end{proposition}

\begin{proposition}[Certified pruning]\label{prop:prune}
For one walk direction, without curvature or mixed probes, with $L\le 3$, let $\hat X_u$ be computed with pruning and $r^w_u$ the probability mass that row $u$ has lost at step $w$. Then
\[
\lVert X_u - \hat X_u\rVert_2 \le \sqrt{5}\,\nTh\,\Big(\textstyle\sum_{w\le L}(r^w_u)^2\Big)^{1/2},
\]
so with a budget $\delta$ the error is at most $\sqrt{5L}\,\nTh\,\delta$. Each mixed probe with lost mass $r_\beta$ adds $4\nTh^2 r_\beta^2$ to the squared bound.
\end{proposition}

The certificate is computed per node from quantities the algorithm already tracks.
It bounds the pruning error in exact arithmetic; it does not cover floating-point or Taylor-evaluation errors, nor downstream prediction accuracy.

\section{Experiments}\label{sec:experiments}

\paragraph{Setup.}
All timings are on one CPU core (Intel Core i9-9900X) unless stated.
Classification uses logistic regression on standardised embeddings over 50 stratified 80/20 splits; paired differences use the Nadeau--Bengio correction for overlapping splits, and are not adjusted for multiple comparisons.
Baselines run with their authors' code and settings: struc2vec \citep{ribeiro2017}, GraphWave \citep{donnat2018}, Role2Vec \citep{ahmed2018}, RESD \citep{resd}, ReFeX \citep{henderson2011} (and a directed variant over in- and out-neighbours), Digraphwave \citep{ceylan2022}, and RWSE \citep{dwivedi2022} with 16 steps.
We add a random-walk heat-kernel characteristic function (RoleWalk's summary applied to $\exp(s(P-I))$ at fixed scales $s\in\{1,2,3\}$, i.e.\ GraphWave without its graph-level quantities), and \emph{local statistics} (degree, neighbour-degree moments, triangles, egonet boundary, clustering, lazy-walk returns at steps 2 and 3), a strong hand-made baseline, and the log-degree alone, which shows how far each task follows activity.
In tables, the best embedding is in bold and the second underlined; the degree baseline is not ranked.

\subsection{Role classification}

\begin{figure}[t]
\centering
\includegraphics[width=\linewidth]{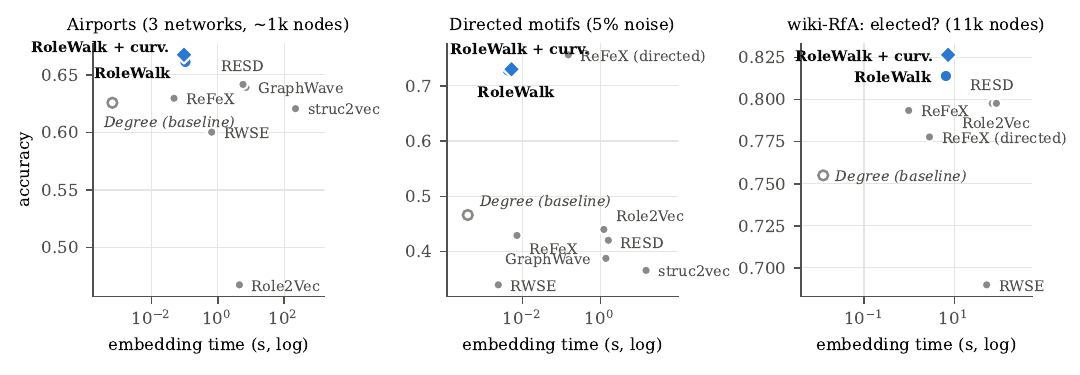}
\caption{Accuracy against embedding time (one core, log scale). Airports: mean of the three networks. Directed motifs: synthetic directed roles with 5\% of edges rewired. wiki-RfA: task A of \cref{tab:rfa}. Hollow marker: degree, a trivial baseline.}
\label{fig:speed}
\end{figure}

\Cref{tab:roles} reports accuracy on three airport networks (labels: activity quartiles), on Reality-call and on Actor (RONE \citep{rone}).
RoleWalk is in the leading group everywhere, with no method dominating; paired over the 50 splits, RoleWalk with curvature is significantly better than GraphWave, the heat-kernel characteristic function, ReFeX, RWSE and local statistics on USA, and no tested comparison shows it significantly worse on these five networks.
Its advantage is cost (\cref{fig:speed}): it embeds the USA network in 0.2\,s, against 8\,s for RESD, 18\,s for GraphWave and 564\,s for struc2vec.
On airports, Brazil and Europe are small (131 and 399 nodes), and paired intervals there have half-widths of about 0.05. The fixed-scale heat-kernel characteristic function, which shares RoleWalk's summary, is close on two airport networks. We leave out the Film network of RONE, where Role2Vec leads by a wide margin (0.743 against 0.581 for RoleWalk with curvature): one class is an independent set in a separate block of node identifiers, and no method does substantially better than the majority class on the others (0.637 at best against 0.623), so we read it as a node-type rather than a role task (supplementary report).

\begin{table}[t]
\centering
\caption{Role classification accuracy (mean over 50 splits; struc2vec: mean $\pm$ std over 5 unseeded runs; ``--'': too slow or out of memory). Time: embedding the USA network (1,190 nodes), one core.}
\label{tab:roles}
\small
\begin{tabular}{lcccccr}
\toprule
& Brazil & Europe & USA & Reality-call & Actor & Time (USA) \\
\midrule
RoleWalk & 0.743 & 0.583 & \second{0.657} & \second{0.860} & \second{0.487} & 0.2\,s \\
RoleWalk + curvature & 0.741 & \second{0.596} & \best{0.666} & \second{0.860} & 0.485 & 0.2\,s \\
struc2vec & 0.705 $\pm$ .022 & 0.561 $\pm$ .020 & 0.596 $\pm$ .004 & -- & -- & 564\,s \\
GraphWave & \best{0.769} & 0.561 & 0.587 & \second{0.860} & 0.481 & 18\,s \\
RESD & 0.728 & 0.544 & 0.653 & 0.857 & 0.476 & 8\,s \\
Role2Vec & 0.476 & 0.366 & 0.562 & 0.844 & 0.399 & 10\,s \\
ReFeX & 0.691 & 0.576 & 0.622 & 0.851 & 0.478 & 0.1\,s \\
Digraphwave & 0.724 & 0.593 & 0.641 & 0.859 & \best{0.488} & 1.1\,s \\
RW heat-kernel CF & \second{0.750} & \best{0.598} & 0.633 & 0.859 & 0.484 & \\
RWSE (16 steps) & 0.711 & 0.572 & 0.517 & 0.855 & 0.363 & 1.7\,s \\
Local statistics & 0.716 & 0.584 & 0.616 & \best{0.862} & 0.486 & \\
\midrule
\emph{Degree (baseline)} & \emph{0.756} & \emph{0.554} & \emph{0.567} & \emph{0.858} & \emph{0.476} & \\
\bottomrule
\end{tabular}
\end{table}

\subsection{Directed roles}

\paragraph{Wikipedia requests for adminship.}
In wiki-RfA \citep{west2014}, 11,377 users cast 188k votes for candidates to adminship.
We predict (A) whether a candidate was elected, and (B) whether a user is a voter, a failed candidate or an admin, from the unsigned, unweighted vote graph alone.
RoleWalk with curvature has the best scores on both tasks (\cref{tab:rfa}); paired over the 50 splits it is significantly better than every other method on A, including Digraphwave ($+0.019$, 95\% interval $[0.008, 0.029]$) and local statistics ($+0.012$ $[0.001, 0.022]$), and on B than every method except local statistics in accuracy ($+0.003$ $[-0.000, 0.006]$).
The mixed probes contribute: without them RoleWalk drops on both tasks, and on the symmetrised graph task B falls from 0.940 to 0.863.
struc2vec and GraphWave do not run at this size.

\begin{table}[t]
\centering
\caption{Real directed roles: wiki-RfA (11,377 users). A: elected among the 3,494 candidates (accuracy). B: voter, failed candidate or admin (accuracy). Time: one core.}
\label{tab:rfa}
\small
\begin{tabular}{lccr}
\toprule
& A: elected & B: 3 classes & Time \\
\midrule
RoleWalk + curvature & \best{0.827} & \best{0.944} & 7\,s \\
RoleWalk & \second{0.814} & 0.940 & 7\,s \\
RoleWalk, no mixed probes & 0.808 & 0.938 & 4\,s \\
Digraphwave & 0.807 & 0.939 & 360\,s \\
ReFeX, directed & 0.778 & 0.930 & 3\,s \\
ReFeX & 0.794 & 0.847 & 1\,s \\
RESD & 0.798 & 0.843 & 68\,s \\
Role2Vec & 0.798 & 0.857 & 82\,s \\
RWSE, out + in & 0.710 & 0.836 & 84\,s \\
Local statistics (in / out) & \second{0.814} & \second{0.941} & 3\,s \\
\midrule
\emph{In/out degree (baseline)} & \emph{0.755} & \emph{0.922} & \\
\bottomrule
\end{tabular}
\end{table}

\paragraph{Network alignment on Digraphwave's protocol.}
Network alignment is where structural embeddings must transfer: a node is matched to the node of another graph whose embedding is closest.
We use the protocol and data of Digraphwave \citep{ceylan2022}: the internal 22,888-node subgraph of the Enron email network is aligned to the full 91,685-node graph, after adding a share $p$ of random edges to their union; each node of the subgraph is matched to its nearest neighbour (euclidean, raw coordinates) among the nodes of the full graph.
This is the benchmark on which Digraphwave reports its largest gains.
RoleWalk runs with its defaults, with no alignment-specific tuning.
We reran Digraphwave with the authors' code and settings: without noise it reproduces their published accuracy at the reported precision (0.398), and on noisy graphs shared with RoleWalk ($p = 0.01, 0.05, 0.1$, two draws each) it agrees closely with their published values (0.321, 0.170, 0.088 against 0.331, 0.170, 0.085), while RoleWalk leads on all six draws, by 0.08, 0.07 and 0.04 on average.

In \cref{tab:align}, RoleWalk, its ablations and the local statistics are evaluated on identical noise draws (5 per level at $p = 0.01, 0.02, 0.03, 0.04, 0.05, 0.075, 0.1$, i.e.\ 35 noisy draws, of which the table shows five levels); the starred rows are published results, not paired with these draws.
Two components carry the result.
The characteristic function is the key block here: with the same out- and in-walks and layers (and no mixed probes, which the top-$k$ ablation lacks), replacing it by the 20 largest probabilities at each step loses 8 points on average, on every one of the 35 noisy draws.
The mixed probes improve every noisy draw, by 0.037 on average.
Two findings temper it.
The layers, scaled by $\nTh$ for raw-distance geometry, dilute nearest-neighbour matching on this task: removing them improves every noisy draw, by 0.020 on average.
And the local statistics, our hand-made baseline, also beat Digraphwave, and their mean is close to RoleWalk's, with no detected difference (RoleWalk minus local statistics $-0.006$ over the 35 noisy draws, paired $t$-test $p = 0.27$): RoleWalk leads up to $p=0.03$, the local statistics beyond.
With email counts as weights, RoleWalk is ahead of Digraphwave at every noise level (0.596 against 0.536 without noise, 0.198 against 0.136 at $p=0.1$), and ReFeX's published results lead at $p=0.1$ (0.227).
On the undirected arXiv graph of the same protocol (aligned to a permuted copy), RoleWalk ties GraphWave and ReFeX without noise (0.784 against 0.785) and leads from $p=0.02$.

\begin{table}[t]
\centering
\caption{Network alignment on the Enron protocol of Digraphwave \citep{ceylan2022} (directed, unweighted): top-1 accuracy of matching the internal subgraph to the full email graph, with a share $p$ of random edges added; mean of 5 noise draws, the same for every row computed here (one deterministic run without noise). $^*$Published results of \citep{ceylan2022}; our rerun of Digraphwave reproduces them (text).}
\label{tab:align}
\small
\begin{tabular}{lcccccc}
\toprule
$p$ & 0 & 0.01 & 0.02 & 0.05 & 0.075 & 0.1 \\
\midrule
RoleWalk & \second{0.487} & \second{0.406} & \second{0.356} & 0.235 & 0.172 & 0.125 \\
\quad characteristic function only (no layers) & \best{0.520} & \best{0.436} & \best{0.379} & \best{0.253} & \second{0.189} & \second{0.142} \\
\quad without mixed probes & 0.442 & 0.363 & 0.316 & 0.198 & 0.140 & 0.097 \\
\quad top-$k$ + layers, without mixed probes & 0.369 & 0.287 & 0.227 & 0.114 & 0.065 & 0.038 \\
\quad layers only & 0.268 & 0.206 & 0.169 & 0.086 & 0.053 & 0.031 \\
Local statistics (in / out) & 0.427 & 0.372 & 0.332 & \second{0.249} & \best{0.207} & \best{0.176} \\
Digraphwave$^*$ & 0.398 & 0.331 & 0.278 & 0.170 & 0.116 & 0.085 \\
ReFeX$^*$ & 0.151 & 0.138 & 0.127 & 0.098 & 0.080 & 0.065 \\
\midrule
\emph{Degree}$^*$ & \emph{0.066} & \emph{0.059} & \emph{0.054} & \emph{0.043} & \emph{0.036} & \emph{0.030} \\
\bottomrule
\end{tabular}
\end{table}

\paragraph{Synthetic directed motifs.}
On directed motifs hung from a directed cycle, where several roles coincide once directions are dropped, RoleWalk recovers every role without noise and has the best retrieval (mAP 0.629) at 5\% rewiring; directed ReFeX is slightly more accurate with logistic regression (0.756 against 0.730) and comparable with gradient boosting (\cref{app:results}).

\subsection{Transfer of centralities across graph sizes}\label{sec:transfer}

Locality (\cref{prop:local}) makes transfer \emph{possible}; whether coordinates actually carry information across graphs is an empirical question.
We fit a gradient-boosting regressor from embeddings to a centrality on one 2,000-node graph and evaluate it (Spearman correlation) on another graph of the same family, four times larger, at the same mean degree.
Six families (Barabási--Albert, powerlaw-cluster, Watts--Strogatz, SBM, random geometric, motifs) with 10 independent pairs of graphs each give 60 pairs.
Only embeddings with fixed coordinates can enter: struc2vec, Role2Vec and RESD learn per-graph coordinates, and ReFeX's feature selection is frozen on the training graph. As a trained inductive model, a GIN \citep{xu2019} (4 layers, 64 hidden units, log-degree input, one configuration, not tuned) is fitted on the training graph to all seven targets at once, with early stopping on a fifth of its nodes.

\begin{table}[t]
\centering
\caption{Centrality transfer to graphs four times larger (Spearman). Table means average the 400 valid graph--target cells of the 60 independent pairs of graphs (20 cells with a constant $k$-core are left out); paired differences in the text first average the valid targets within each pair, then the pairs. $^\dagger$The clustering coefficient is itself one of the features of the local statistics and of ReFeX. $^\ddagger$Without its $1/n$ factor. GIN: trained on the training graph (log-degree input).}
\label{tab:centrality}
\small
\setlength{\tabcolsep}{4pt}
\begin{tabular}{lcccccccc}
\toprule
& RoleWalk & Local stats & ReFeX (frozen) & RW heat CF & GIN & RWSE & GraphWave$^\ddagger$ & \emph{Degree} \\
\midrule
Betweenness & \best{0.864} & \second{0.852} & 0.845 & 0.849 & 0.778 & 0.412 & 0.659 & \emph{0.623} \\
Closeness & \best{0.797} & 0.716 & 0.725 & \second{0.738} & 0.695 & 0.464 & 0.529 & \emph{0.407} \\
Harmonic & \best{0.861} & 0.778 & 0.788 & \second{0.806} & 0.756 & 0.517 & 0.597 & \emph{0.460} \\
PageRank & \best{0.989} & 0.984 & 0.973 & \second{0.986} & 0.982 & 0.741 & 0.611 & \emph{0.929} \\
Clustering & 0.805 & \best{0.844}$^\dagger$ & \second{0.813}$^\dagger$ & 0.785 & 0.387 & 0.723 & 0.347 & \emph{0.201} \\
$k$-core & \best{0.651} & \second{0.650} & 0.631 & 0.647 & 0.604 & 0.494 & 0.594 & \emph{0.614} \\
Eigenvector & 0.673 & 0.661 & \second{0.700} & 0.586 & \best{0.736} & 0.358 & 0.415 & \emph{0.463} \\
\midrule
Mean & \best{0.813} & \second{0.790} & \second{0.790} & 0.777 & 0.711 & 0.532 & 0.533 & \emph{0.524} \\
\bottomrule
\end{tabular}
\end{table}

RoleWalk has the best mean (\cref{tab:centrality}).
Paired over the 60 pairs of graphs, it is ahead of local statistics by $+0.022$ $[0.016,0.029]$ (80\% of pairs won), of frozen ReFeX by $+0.024$ $[0.018,0.029]$ of the random-walk heat-kernel characteristic function by $+0.038$ $[0.030,0.046]$ (98\% of pairs), and of the trained GIN by $+0.102$ $[0.087, 0.116]$, on every pair; the GIN is best only on eigenvector centrality. Feeding RoleWalk to the GIN as input lowers its mean to 0.675, and RWSE to 0.496: on one training graph, the gradient-boosted regressor on fixed features transfers better than message passing.
On closeness and harmonic centrality it wins every pair against local statistics and ReFeX.
GraphWave and RWSE lose more with size: going from test graphs of the training size to four times that size (2 seeds at equal size, so descriptively), RoleWalk's mean drops by 0.04, GraphWave's by 0.10 and RWSE's by 0.19.

\paragraph{What carries the transfer?}
Splitting RoleWalk into its two blocks shows that the law of the distance does most of the work: the layers alone reach 0.807 and the characteristic function alone 0.785, and the full embedding adds a small but consistent gain over the layers ($+0.007$ $[0.004,0.009]$). The return probabilities alone of the same four-step lazy walk (a lazy RWSE) reach 0.753, against 0.807 for all the layers; the 0.532 of the RWSE column uses a different walk and 16 steps.
Replacing the characteristic function by the 20 largest probabilities at each step, with the same layers and the same dimension, gives 0.812 --- a tie.
What transfers, then, is the distance law of a short walk, which is the main new ingredient of RoleWalk relative to GraphWave-style signatures.

\paragraph{Beyond size.}
When the test graph instead comes from the same generator with one parameter changed (mean degree, clustering, rewiring, community strength, density), RoleWalk has the highest mean among the embeddings, though its advantage over frozen ReFeX is not established, and the best means on closeness and harmonic centrality; local statistics are better on average, mostly on the clustering target, which is one of their features (\cref{app:results}). Here the characteristic function does help: RoleWalk is ahead of the dimension-matched top-$k$ ablation ($+0.024$ $[0.003, 0.044]$).

\subsection{Scale and certified pruning}

\begin{table}[t]
\centering
\caption{Embedding time on one core (defaults: $\tol=10^{-4}$ without budget; one run after compiling the kernel; including the conversion of the networkx graph).}
\label{tab:scale}
\small
\begin{tabular}{lrrrrr}
\toprule
Nodes & 10k & 30k & 100k & 300k & 1M \\
\midrule
Barabási--Albert ($m=3$) & 0.5\,s & 1.8\,s & 8.9\,s & 43\,s & 193\,s \\
\quad + curvature & 0.6\,s & 1.9\,s & 11\,s & 52\,s & 247\,s \\
Watts--Strogatz ($k=6$) & 0.1\,s & 0.5\,s & 1.8\,s & 5.1\,s & 18\,s \\
\bottomrule
\end{tabular}
\end{table}

RoleWalk embeds a million-node graph in 18\,s to 3.2\,min on one core, with at most 2.5\,GB of peak memory, most of which holds the networkx graph (\cref{tab:scale}).
Among the baselines, only ReFeX also reaches a million nodes (31\,s, faster than RoleWalk); RESD takes over an hour; at 100k nodes, 16-step RWSE exceeds 16\,GB, Role2Vec 10\,GB, and the heat-kernel characteristic function and Digraphwave (on CPU) an hour, and GraphWave and struc2vec do not run on wiki-RfA (11k nodes).

Threshold pruning reduces running time in our measurements but can produce large errors on the worst nodes: on the Barabási--Albert graph with a million nodes (a separate run, 173\,s), pruning at $\tol=10^{-4}$ leaves 69\% of the rows missing more than 1\% of their mass, and the worst certificate is 366.
With a 1\% mass budget, the run takes 227\,s and every certificate is below 3.9, against embedding norms of a few hundred; exact computation at this scale was not timed.
On 50k-node graphs, the observed error never exceeds 0.64 of the certificate, and a 1\% budget brings the worst-node relative error from 22\% to 0.7\% (\cref{app:results}).
A user can thus check the worst case of a pruned run instead of trusting the median.

\subsection{Expressiveness and use as a structural encoding}

On BREC \citep{wang2023} (400 pairs of non-isomorphic graphs that 1-WL cannot separate), RoleWalk with 3 steps separates 207 pairs and with 5 steps 213, against 214 for RWSE with 16 steps; as predicted by \cref{prop:expr}(c), neither separates any distance-regular pair.
As a structural encoding in GPS \citep{rampasek2022} on ZINC, adding RoleWalk with curvature to RWSE lowers the test error on each of 5 seeds (0.072 against 0.077 for our RWSE reproduction), but the paired interval just includes zero and our reproduction is weaker than the published 0.070: we read this as preliminary.

\section{Beyond square adjacency}

Nothing in RoleWalk requires the walk to stay on one set of nodes.
The characteristic function is defined for any row of a row-stochastic matrix, square or not, and the products that build the walk only need to compose: with $B$ the row-normalised incidence from nodes to edges, $B B'$ walks node $\to$ edge $\to$ node, and incidences between edges and faces walk across the ranks of a cell complex or a hypergraph; the distance law becomes the number of steps between cells.
The mixed probes are already a case of this: $H_+H_-$ composes two different operators.
Diffusion signatures need a square operator to exponentiate; walks across ranks can be written that way too, as one block operator on the disjoint union of all cells, so this is a convenience rather than a capability that diffusion lacks: RoleWalk composes the rectangular steps directly, with no spectral computation and the same locality.
Walks between cells of different dimensions --- atoms, bonds and rings in molecules, for instance --- are a direct extension of RoleWalk, which we leave to future work.

\section{Limitations}

\begin{itemize}
\item \textbf{Provable blind spots.} Walk statistics cannot separate distance-regular graphs with equal intersection arrays, most CFI pairs, or graphs whose neighbourhoods coincide; without curvature, some nodes separated by 1-WL are confused.
\item \textbf{Simple baselines remain strong.} ReFeX is faster and slightly more accurate on noisy synthetic directed motifs with a linear classifier; hand-made local statistics transfer better than RoleWalk when generator parameters change, and match it on average on Enron alignment; on a second real directed setting, food webs with LLM-generated proxy trophic labels, RoleWalk, local statistics and Digraphwave are not distinguishable (supplementary report).
\item \textbf{Pruning.} Without a budget the certificate is only a posteriori and large on hubs; the budget costs time on hub-heavy graphs.
\item \textbf{Labels.} Airport labels are activity quartiles and favour degree-like features; in transfer between airport networks, local statistics and the degree alone beat every embedding. Real labelled directed role datasets remain scarce.
\end{itemize}

\section{Conclusion}

RoleWalk describes each node by two laws of a short lazy walk --- how concentrated it is and how far it goes --- with fixed functions that need no training and no graph-level quantity.
This makes its coordinates comparable across graphs, which is what alignment and transfer need, and lets it read directions through walks that change direction.
It is competitive with learned role embeddings at a small fraction of their cost, has the highest scores on wiki-RfA, exceeds Digraphwave on its own alignment benchmark, and scales to a million nodes on one core with certified pruning.
The ablations show that its components serve different tasks: the law of the walker's distance carries centralities across graph sizes, a simple ingredient that other structural encodings could adopt, while the characteristic function and the mixed probes carry network alignment.

\bibliographystyle{abbrvnat}
\bibliography{refs}

\begin{thebibliography}{27}
\providecommand{\natexlab}[1]{#1}
\providecommand{\url}[1]{\texttt{#1}}
\expandafter\ifx\csname urlstyle\endcsname\relax
  \providecommand{\doi}[1]{doi: #1}\else
  \providecommand{\doi}{doi: \begingroup \urlstyle{rm}\Url}\fi

\bibitem[Ahmed et~al.(2018)Ahmed, Rossi, Lee, Willke, Zhou, Kong, and
  Eldardiry]{ahmed2018}
N.~K. Ahmed, R.~Rossi, J.~B. Lee, T.~L. Willke, R.~Zhou, X.~Kong, and
  H.~Eldardiry.
\newblock Learning role-based graph embeddings.
\newblock \emph{arXiv:1802.02896}, 2018.

\bibitem[Brouwer et~al.(1989)Brouwer, Cohen, and Neumaier]{brouwer1989}
A.~E. Brouwer, A.~M. Cohen, and A.~Neumaier.
\newblock \emph{Distance-Regular Graphs}.
\newblock Springer, 1989.

\bibitem[Ceylan et~al.(2022)Ceylan, Ghoorchian, and Kragic]{ceylan2022}
C.~Ceylan, K.~Ghoorchian, and D.~Kragic.
\newblock Digraphwave: Scalable extraction of structural node embeddings via
  diffusion on directed graphs.
\newblock \emph{arXiv:2207.10149}, 2022.

\bibitem[Donnat et~al.(2018)Donnat, Zitnik, Hallac, and Leskovec]{donnat2018}
C.~Donnat, M.~Zitnik, D.~Hallac, and J.~Leskovec.
\newblock Learning structural node embeddings via diffusion wavelets.
\newblock In \emph{KDD}, 2018.

\bibitem[Dwivedi et~al.(2022)Dwivedi, Luu, Laurent, Bengio, and
  Bresson]{dwivedi2022}
V.~P. Dwivedi, A.~T. Luu, T.~Laurent, Y.~Bengio, and X.~Bresson.
\newblock Graph neural networks with learnable structural and positional
  representations.
\newblock In \emph{ICLR}, 2022.

\bibitem[Gretton et~al.(2012)Gretton, Borgwardt, Rasch, Sch{\"o}lkopf, and
  Smola]{gretton2012}
A.~Gretton, K.~M. Borgwardt, M.~J. Rasch, B.~Sch{\"o}lkopf, and A.~Smola.
\newblock A kernel two-sample test.
\newblock \emph{JMLR}, 13, 2012.

\bibitem[Heimann et~al.(2018)Heimann, Shen, Safavi, and Koutra]{heimann2018}
M.~Heimann, H.~Shen, T.~Safavi, and D.~Koutra.
\newblock {REGAL}: Representation learning-based graph alignment.
\newblock In \emph{Proceedings of the 27th ACM International Conference on
  Information and Knowledge Management (CIKM)}, pages 117--126, 2018.

\bibitem[Henderson et~al.(2011)Henderson, Gallagher, Li, Akoglu, Eliassi-Rad,
  Tong, and Faloutsos]{henderson2011}
K.~Henderson, B.~Gallagher, L.~Li, L.~Akoglu, T.~Eliassi-Rad, H.~Tong, and
  C.~Faloutsos.
\newblock It's who you know: graph mining using recursive structural features.
\newblock In \emph{KDD}, 2011.

\bibitem[Henderson et~al.(2012)Henderson, Gallagher, Eliassi-Rad, Tong, Basu,
  Akoglu, Koutra, Faloutsos, and Li]{henderson2012}
K.~Henderson, B.~Gallagher, T.~Eliassi-Rad, H.~Tong, S.~Basu, L.~Akoglu,
  D.~Koutra, C.~Faloutsos, and L.~Li.
\newblock {RolX}: structural role extraction and mining in large graphs.
\newblock In \emph{KDD}, 2012.

\bibitem[Ivanov and Burnaev(2018)]{ivanov2018anonymous}
S.~Ivanov and E.~Burnaev.
\newblock Anonymous walk embeddings.
\newblock In \emph{ICML}, pages 2186--2195, 2018.

\bibitem[Jiao et~al.(2021)Jiao, Guo, Pan, Zhang, and Pei]{rone}
P.~Jiao, X.~Guo, T.~Pan, W.~Zhang, and Y.~Pei.
\newblock A survey on role-oriented network embedding.
\newblock \emph{arXiv preprint arXiv:2107.08379}, 2021.

\bibitem[Jost and Liu(2014)]{jost2014}
J.~Jost and S.~Liu.
\newblock Ollivier's {R}icci curvature, local clustering and
  curvature-dimension inequalities on graphs.
\newblock \emph{Discrete \& Computational Geometry}, 51, 2014.

\bibitem[Kessler(1963)]{kessler1963}
M.~M. Kessler.
\newblock Bibliographic coupling between scientific papers.
\newblock \emph{American Documentation}, 14\penalty0 (1):\penalty0 10--25,
  1963.

\bibitem[Li et~al.(2020)Li, Wang, Wang, and Leskovec]{li2020distance}
P.~Li, Y.~Wang, H.~Wang, and J.~Leskovec.
\newblock Distance encoding: Design provably more powerful neural networks for
  graph representation learning.
\newblock In \emph{NeurIPS}, 2020.

\bibitem[Ma et~al.(2019)Ma, Qin, Qiu, Zheng, and Wang]{ma2019riwalk}
X.~Ma, G.~Qin, Z.~Qiu, M.~Zheng, and Z.~Wang.
\newblock {RiWalk}: Fast structural node embedding via role identification.
\newblock In \emph{ICDM}, pages 478--487, 2019.

\bibitem[Morris et~al.(2019)Morris, Ritzert, Fey, Hamilton, Lenssen, Rattan,
  and Grohe]{morris2019}
C.~Morris, M.~Ritzert, M.~Fey, W.~L. Hamilton, J.~E. Lenssen, G.~Rattan, and
  M.~Grohe.
\newblock Weisfeiler and {L}eman go neural: higher-order graph neural networks.
\newblock In \emph{AAAI}, 2019.

\bibitem[Ollivier(2009)]{ollivier2009}
Y.~Ollivier.
\newblock Ricci curvature of {M}arkov chains on metric spaces.
\newblock \emph{Journal of Functional Analysis}, 256\penalty0 (3), 2009.

\bibitem[Ramp{\'a}{\v{s}}ek et~al.(2022)Ramp{\'a}{\v{s}}ek, Galkin, Dwivedi,
  Luu, Wolf, and Beaini]{rampasek2022}
L.~Ramp{\'a}{\v{s}}ek, M.~Galkin, V.~P. Dwivedi, A.~T. Luu, G.~Wolf, and
  D.~Beaini.
\newblock Recipe for a general, powerful, scalable graph transformer.
\newblock In \emph{NeurIPS}, 2022.

\bibitem[Ribeiro et~al.(2017)Ribeiro, Saverese, and Figueiredo]{ribeiro2017}
L.~F.~R. Ribeiro, P.~H.~P. Saverese, and D.~R. Figueiredo.
\newblock struc2vec: learning node representations from structural identity.
\newblock In \emph{KDD}, 2017.

\bibitem[Rozemberczki and Sarkar(2020)]{rozemberczki2020feather}
B.~Rozemberczki and R.~Sarkar.
\newblock Characteristic functions on graphs: Birds of a feather, from
  statistical descriptors to parametric models.
\newblock In \emph{CIKM}, pages 1325--1334, 2020.

\bibitem[Small(1973)]{small1973}
H.~Small.
\newblock Co-citation in the scientific literature: a new measure of the
  relationship between two documents.
\newblock \emph{Journal of the American Society for Information Science},
  24\penalty0 (4):\penalty0 265--269, 1973.

\bibitem[Steenfatt et~al.(2019)Steenfatt, Nikolentzos, Vazirgiannis, and
  Zhao]{steenfatt2018}
N.~Steenfatt, G.~Nikolentzos, M.~Vazirgiannis, and Q.~Zhao.
\newblock Learning structural node representations on directed graphs.
\newblock In \emph{Complex Networks and Their Applications VII}, volume 813 of
  \emph{Studies in Computational Intelligence}, pages 132--144. Springer, 2019.
\newblock \doi{10.1007/978-3-030-05414-4_11}.

\bibitem[Wang and Zhang(2023)]{wang2023}
Y.~Wang and M.~Zhang.
\newblock Towards better evaluation of {GNN} expressiveness with {BREC}
  dataset.
\newblock \emph{arXiv:2304.07702}, 2023.

\bibitem[Weisfeiler and Leman(1968)]{weisfeiler1968}
B.~Weisfeiler and A.~Leman.
\newblock The reduction of a graph to canonical form and the algebra which
  appears therein.
\newblock \emph{NTI, Series 2}, 1968.

\bibitem[West et~al.(2014)West, Paskov, Leskovec, and Potts]{west2014}
R.~West, H.~S. Paskov, J.~Leskovec, and C.~Potts.
\newblock Exploiting social network structure for person-to-person sentiment
  analysis.
\newblock \emph{TACL}, 2, 2014.

\bibitem[Xu et~al.(2019)Xu, Hu, Leskovec, and Jegelka]{xu2019}
K.~Xu, W.~Hu, J.~Leskovec, and S.~Jegelka.
\newblock How powerful are graph neural networks?
\newblock In \emph{ICLR}, 2019.

\bibitem[Zhang et~al.(2021)Zhang, Guo, Wang, Tian, Pan, and Jiao]{resd}
W.~Zhang, X.~Guo, W.~Wang, Q.~Tian, L.~Pan, and P.~Jiao.
\newblock Role-based network embedding via structural features reconstruction
  with degree-regularized constraint.
\newblock \emph{Knowledge-Based Systems}, 218:\penalty0 106872, 2021.
\newblock \doi{10.1016/j.knosys.2021.106872}.

\end{thebibliography}

\clearpage
\appendix
\section{Proofs}\label{app:proofs}

Throughout, $L$ is the number of steps, graphs have finite non-negative weights and no self-loops, layer distances are shortest paths in the graph of positive-weight edges, in the direction of the walk; locality balls use the underlying undirected graph on directed graphs and with curvature; $m_x$ and $\kappa$ are defined on the simple unweighted symmetrisation, and $\tol=0$.

\begin{proof}[Proof of \cref{prop:local}]
$p^w_{uj} = \sum_{u=x_0,\dots,x_w=j}\prod_t H_{x_tx_{t+1}}$ runs over walks of length $w\le L$ from $u$.
Their nodes lie within distance $w$, and $H_{xy}$ needs the edges at $x$ (and $s_x$, $d_x$), with $x$ within distance $w-1$: all lie in $B_L(u)$.
An isomorphism of rooted balls maps these walks bijectively and preserves their weights, so it maps $p^w_u$ to the corresponding row entry by entry; \eqref{eq:phi} is a sum over $j$, hence equal. Distances up to $L$ are read inside $B_L(u)$.
With curvature, $\kappa_j$ for $j$ within distance $L$ needs the closed neighbourhoods of $j$ and of its neighbours, inside $B_{L+2}(u)$.
A mixed probe goes from $u$ to a neighbour $x$ in one direction, then to a neighbour of $x$ in the other: it needs the edges at $u$ and at $x$, within undirected distance 2 of $u$.
\end{proof}

\begin{lemma}[Injectivity]\label{lem:inj}
Let $P,Q$ be finite multisets of reals in $(0,1]$. If $\sum_{p\in P}f_\theta(p) = \sum_{q\in Q}f_\theta(q)$ for all $\theta$ in a set with an accumulation point, then $P=Q$. For a finite grid, given any countable family of pairs $P\ne Q$, Lebesgue-almost every $\Th\in\R^m$ separates all of them.
\end{lemma}
\begin{proof}
$F(\theta) = \sum_P e^{i\theta p} - \sum_Q e^{i\theta q} - (|P|-|Q|)$ is entire and vanishes on a set with an accumulation point, hence everywhere. Exponentials with distinct frequencies are linearly independent, so each $\lambda>0$ has the same multiplicity in $P$ and $Q$. For the second claim, the real zeros of each nonzero entire $F$ are countable, and a countable union of countable sets is null.
\end{proof}

\begin{proof}[Proof of \cref{prop:mmd}]
$X_u = \sum\Psi$ over the atoms of $\mu_u$, with $\Psi(w,a,\ell) = e_w\otimes(\psi(a)\oplus\nTh\,a\,e_\ell)$ and $\psi(a) = (f_\theta(a))_{\theta\in\Th}$ split into real and imaginary parts. So $X_u$ is the kernel mean embedding of $\mu_u$ and $\lVert X_u-X_v\rVert^2$ is the MMD for $K = \operatorname{Re}\langle\Psi,\Psi'\rangle$, positive semidefinite as an inner product. Orthonormal blocks give $\delta_{ww'}$ and $\delta_{\ell\ell'}$; expanding $\operatorname{Re}(e^{i\theta a}-1)\overline{(e^{i\theta b}-1)}$ gives the first term.
For the first-order statement, write the real part as $(\cos\theta a-1)(\cos\theta b-1)+\sin\theta a\sin\theta b$ and use $|\cos x-1|\le x^2/2$ and $|\sin x - x|\le x^3/6$: with $t = \max(\Th)\max(a,b)$,
$\big|\sum_\theta[\dots] - \nTh^2ab\big| \le \nTh^2 ab\, t^2(7/12 + t^2/36)$, and $\nTh^2=\sum_\theta\theta^2$ is the coefficient of $ab$.
\end{proof}

\begin{proof}[Proof of \cref{prop:expr}]
(a) On a $d$-regular graph $H = (I+A)/(d+1)$. At step 2, the mass on the neighbours of $u$ is $\sum_{j\sim u}(I+2A+A^2)_{uj}/(d+1)^2$, with $\sum_{j\sim u}A_{uj}=d$ and $\sum_{j\sim u}(A^2)_{uj}=2t_u$. 1-WL gives every node of two $d$-regular graphs of the same size the same colour.

(b) Take a clique $K_5$ on $\{r,a,b,c,d\}$ and ten further nodes, adjacent to $b,c,d$ in $G$, and five to $a,b$ and five to $c,d$ in $G'$. The neighbour degrees of $r$ are $\{4,14,14,14\}$ in $G$ and $\{9,9,9,9\}$ in $G'$, so two rounds of 1-WL separate $r$. Walks from $r$ agree: at step 1 each clique node has probability $1/5$; at step 2 each clique node has $3/25$ and each outer node $1/25$ in both graphs, with the same layers. For $L=3$, add a leaf $z$ at $r$ and root at $z$: three rounds separate $z$, while step 3 puts $2/9$ on $z$, $23/90$ on $r$, $4/45$ on each of $a,b,c,d$ and $1/60$ on each outer node in both graphs (exact rational computation, released with the code).

(c) Let $A_i$ be the distance-$i$ matrices. Distance-regularity gives $AA_i = b_{i-1}A_{i-1}+a_iA_i+c_{i+1}A_{i+1}$ with coefficients from the intersection array \citep{brouwer1989}, so $(I+A)^w = \sum_i\alpha_{w,i}A_i$ with $\alpha_{w,i}$ depending only on the array, and $p^w_{uj} = \alpha_{w,d(u,j)}/(k+1)^w$. Each row has value $\alpha_{w,i}/(k+1)^w$ with multiplicity $k_i$, the number of nodes at distance $i$, also fixed by the array; so $\phi^w_u$ and $s^w_u$ are the same at all nodes of both graphs. For an edge $xy$, $\kappa(x,y) = (a_1+2)/(k+1)$, a constant.
\end{proof}

\begin{proof}[Proof of \cref{prop:mixed}]
Out-walk: $u$ keeps $1/3$ and sends $1/3$ to each sink, where it stays; the layers agree. In-walk: $u$ has no in-edge and stays. The symmetrised graphs are a path and a 4-cycle, where every node has $\kappa=2/3$. In $H_-$, a sink with in-neighbours $N$ spreads uniformly over itself and $N$: $1/2$ on itself and $u$ in the first graph, $1/3$ on itself, $u$ and the other source in the second. Hence $e_u^\top H_+H_- = \frac13(e_u+\sum_{\text{sinks } s}e_s^\top H_-)$, which gives the two rows.
\end{proof}

\begin{proof}[Proof of \cref{prop:prune}]
Propagation and deletion keep entries non-negative, so the pruned row $q^w_u$ satisfies $0\le q^w_u\le p^w_u$ and $\lVert p^w_u - q^w_u\rVert_1 = r^w_u$. For each $\theta$, $|\sum_j f_\theta(p_j)-f_\theta(q_j)|\le\theta r^w_u$, and the first-order correction for the dropped mass adds at most $\theta r^w_u$: the characteristic-function block moves by at most $2\nTh r^w_u$. Layers use exact distances for $L\le3$, so each layer loses a non-negative mass and in total $r^w_u$: the scaled layer block moves by at most $\nTh r^w_u$. Summing squares over blocks and steps gives the bound. A budget enforces $r^w_u\le\delta$, since a stochastic matrix preserves the missing mass and a row only drops entries while it stays below $\delta$. A mixed probe is a characteristic function without layers, hence $2\nTh r_\beta$.
\end{proof}

\begin{proposition}[Relation to GraphWave]\label{prop:graphwave}
GraphWave embeds $u$ by $\chi_u(t) = \frac1n\sum_j e^{it\Psi_{ju}}$ with $\Psi$ the heat kernel at a scale chosen from each graph's spectrum. Then $n(\chi_u(t)-1) = \sum_j f_t(\Psi_{ju})$: \eqref{eq:phi} with the heat kernel in place of $H^w$. Since $\chi_u$ depends on $n$ and on the scale, the same rooted neighbourhood can get different coordinates in graphs of different sizes or spectra; removing the $1/n$ lifts GraphWave's airport transfer from chance to 0.455 (RoleWalk: 0.488).
\end{proposition}

\begin{proposition}[Relation to RWSE]
The distance-0 layer $s^w_u(0) = (H^w)_{uu}$ is the return probability of the lazy walk. On a simple, undirected, unweighted $d$-regular graph with $d>0$, $\{(H^w)_{uu}\}_{w\le L}$ and the simple-walk RWSE $\{(P^k)_{uu}\}_{k\le L}$ determine each other, since $(H^w)_{uu} = (d+1)^{-w}\sum_{k}\binom{w}{k}d^k(P^k)_{uu}$ is triangular with nonzero diagonal.
\end{proposition}

\section{Additional results}\label{app:results}

\begin{table}[!ht]
\centering
\caption{Synthetic directed roles (motifs hung from a directed cycle; 3 seeds; noise: share of edges rewired). Accuracy: logistic regression; ARI: $k$-means; mAP: retrieval.}
\small
\begin{tabular}{lccccc}
\toprule
& Acc. 0\% & Acc. 5\% & Acc. 10\% & ARI 5\% & mAP 5\% \\
\midrule
RoleWalk + curvature & \best{1.000} & \second{0.730} & \second{0.593} & \second{0.396} & \best{0.629} \\
RoleWalk & \best{1.000} & 0.728 & 0.588 & 0.373 & \second{0.615} \\
ReFeX, directed & \best{1.000} & \best{0.756} & \best{0.638} & 0.355 & 0.607 \\
Digraphwave & \best{1.000} & 0.717 & 0.575 & 0.292 & 0.539 \\
Local statistics (in / out) & 0.863 & 0.675 & 0.584 & \best{0.409} & 0.609 \\
GraphWave & 0.586 & 0.388 & 0.383 & 0.317 & 0.416 \\
struc2vec & 0.510 & 0.366 & 0.277 & 0.141 & 0.260 \\
\midrule
\emph{In/out degree} & \emph{0.516} & \emph{0.466} & \emph{0.405} & \emph{0.310} & \emph{0.349} \\
\bottomrule
\end{tabular}
\end{table}
With gradient boosting instead of logistic regression, RoleWalk with curvature reaches 1.000 / 0.744 / 0.586 and directed ReFeX 1.000 / 0.739 / 0.578: the two are comparable on these motifs.

\begin{table}[!ht]
\centering
\caption{Centrality transfer under a change of generator parameter (Barabási--Albert $m=3\to5$, powerlaw-cluster triangle probability $0.3\to0.6$, Watts--Strogatz rewiring $0.1\to0.3$, SBM within-block probabilities $\times2$, random geometric radius $0.04\to0.06$; 50 pairs). $^\dagger$A feature of these methods.}
\small
\begin{tabular}{lccccc}
\toprule
& RoleWalk & Local stats & ReFeX (frozen) & RW heat CF & GraphWave \\
\midrule
Betweenness & 0.614 & \best{0.742} & \second{0.701} & 0.614 & 0.619 \\
Closeness & \best{0.720} & \second{0.618} & 0.597 & 0.599 & 0.490 \\
Harmonic & \best{0.750} & 0.657 & 0.624 & \second{0.663} & 0.567 \\
PageRank & 0.893 & \best{0.923} & 0.831 & 0.677 & \second{0.878} \\
Clustering & 0.639 & \best{0.959}$^\dagger$ & \second{0.759}$^\dagger$ & 0.539 & 0.040 \\
$k$-core & $-0.017$ & \second{0.134} & 0.018 & $-0.065$ & \best{0.235} \\
Eigenvector & \best{0.612} & \second{0.605} & 0.554 & 0.533 & 0.513 \\
\midrule
Mean & \second{0.639} & \best{0.695} & 0.618 & 0.543 & 0.492 \\
\bottomrule
\end{tabular}
\end{table}
Without the clustering target, RoleWalk and local statistics are not distinguishable (paired difference $-0.008$, interval $[-0.033, 0.017]$), and RoleWalk is ahead of frozen ReFeX ($+0.049$, interval $[0.020, 0.078]$).

\begin{table}[!ht]
\centering
\caption{Certified pruning on 50k-node graphs and wiki-RfA (one core): time, relative error of each node's embedding against exact computation (median and worst node), and largest per-node certificate.}
\small
\begin{tabular}{llrrrr}
\toprule
Graph & Setting & Time & Median & Worst & Cert. max \\
\midrule
Barabási--Albert 50k & exact & 3.8\,s & 0.0\% & 0.0\% & -- \\
 & $\tol=10^{-4}$ & 3.0\,s & 1.3\% & 22.4\% & 188 \\
 & $\tol=10^{-4}$, $\delta=0.01$ & 3.6\,s & 0.0\% & 0.7\% & 4 \\
Directed scale-free 50k & exact & 38.2\,s & 0.0\% & 0.0\% & -- \\
 & $\tol=10^{-4}$ & 7.8\,s & 0.6\% & 44.9\% & 414 \\
 & $\tol=10^{-4}$, $\delta=0.01$ & 43.1\,s & 0.0\% & 0.6\% & 3 \\
wiki-RfA & exact & 4.1\,s & 0.0\% & 0.0\% & -- \\
 & $\tol=10^{-4}$ & 3.6\,s & 1.6\% & 6.7\% & 77 \\
 & $\tol=10^{-4}$, $\delta=0.01$ & 4.3\,s & 0.0\% & 0.5\% & 5 \\
\bottomrule
\end{tabular}
\end{table}

\end{document}